\documentclass[runningheads]{llncs}
\usepackage{graphicx}
\usepackage[algoruled,linesnumbered]{algorithm2e}
\usepackage{amsfonts}       % blackboard math symbols
\usepackage{amsmath}
\usepackage{nicefrac}       % compact symbols for 1/2, etc.
\usepackage{subcaption}
\usepackage{booktabs}
\usepackage{multirow} 
\usepackage{wrapfig}
\usepackage{xcolor}
\usepackage{hyperref}

\newcommand{\argmin}{\mathop{\mathrm{argmin}}}
\newcommand{\argmax}{\mathop{\mathrm{argmax}}}
\newcommand{\mfkd}{\mbox{MF-KD}}

\begin{document}
\title{Multi-fidelity Neural Architecture Search with Knowledge Distillation}

\author{
Ilya Trofimov$^{1}$
\and
Nikita Klyuchnikov$^{1}$
\and
Mikhail Salnikov$^{1}$
\and\\
Alexander Filippov$^{2}$
\and
Evgeny Burnaev$^{1}$}

\authorrunning{I. Trofimov et al.}
\institute{Skolkovo Institute of Science and Technology, Moscow, Russia \and
Huawei Noah's Ark Lab, Moscow, Russia}
\maketitle              % typeset the header of the contribution
\begin{abstract}
  Neural architecture search (NAS) targets at finding the optimal architecture of a neural network for a problem or a family of problems. Evaluations of neural architectures are very time-consuming. One of the possible ways to mitigate this issue is to use low-fidelity evaluations, namely training on a part of a dataset, fewer epochs, with fewer channels, etc. In this paper, we propose a bayesian multi-fidelity method for neural architecture search: \mfkd{}. The method relies on a new approach to low-fidelity evaluations of neural architectures by training for a few epochs using a knowledge distillation. Knowledge distillation adds to a loss function a term forcing a network to mimic some teacher network. We carry out experiments on CIFAR-10, CIFAR-100, and ImageNet-16-120.
  We show that training for a few epochs with such a modified loss function leads to a better selection of neural architectures than training for a few epochs with a logistic loss.
  %The proposed low-fidelity evaluations were incorporated into a bayesian multi-fidelity search algorithm (co-kriging) that outperformed the search based on high-fidelity evaluations only (training until convergence) and few epochs with conventional logistic loss.  
  The proposed method outperforms several state-of-the-art baselines. 
  %the search based on high-fidelity evaluations only (training until convergence) and few epochs with conventional logistic loss.  

\end{abstract}
\section{Introduction}

\begin{figure}[t]
\centering
\includegraphics[angle=0,width=\textwidth]{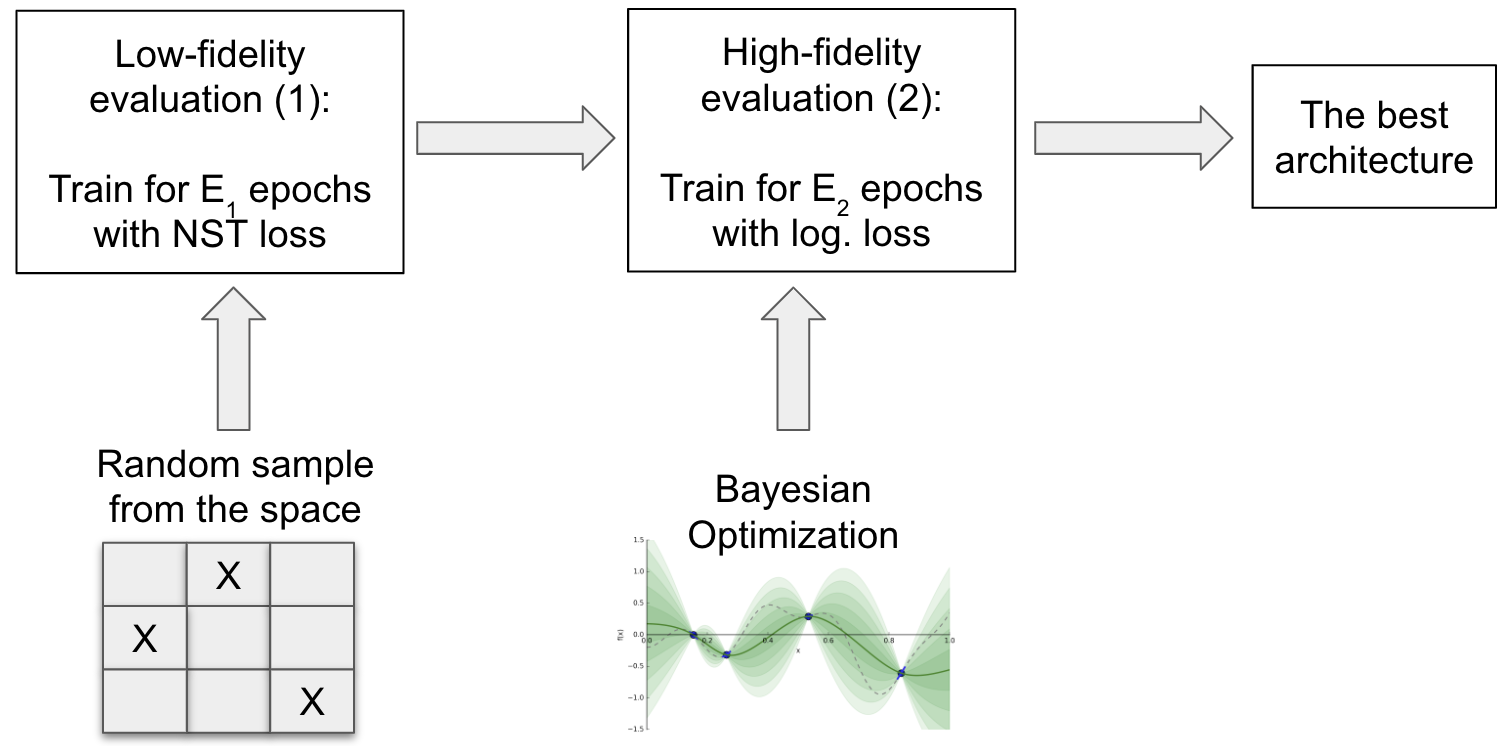}
\caption{High-level description of the proposed \mfkd{}  method.}
\label{fig:flowchart}
\end{figure}

Deep learning is state of the art in the majority of ML problems: computer vision, speech recognition, machine translation, etc. The progress in this area mostly comes from discovering new architectures of neural networks, which is usually performed by human experts. This motivates a new direction of research -- \textit{neural architecture search (NAS)} -- developing algorithms for finding new well-performing architectures of neural networks.
Existing approaches could be broadly divided into two groups.

\textbf{Black-box optimization}. Given a discrete search space $\mathcal{A}$ of all the architectures and a performance function $f(\cdot)$ of an architecture, like testing accuracy, these approaches aim to solve $\argmax_{a \in \mathcal{A}} f(a) $ via black-box optimization.
One of the first proposed approaches of this kind \cite{zoph2016neural,zoph2018learning} treated architecture design (layer by layer) as a sequential decision process and the performance $f(a)$ was a reward. The optimization was done by reinforcement learning. 
Classical methods like Gaussian process-based bayesian optimization with a particular kernel \cite{kandasamy2018neural} and evolutionary optimization \cite{real2019regularized} also could be applied.
Some methods \cite{white2019bananas,shi2019multi} use performance predictors together with bayesian uncertainty estimation.
The black-box optimization methods share the same drawback -- they require a large number of architecture evaluations and significant computational resources. 

\textbf{One-shot NAS}. Another line of research goes beyond black-box optimization and utilizes the structure and the learning algorithm of a neural network. The architecture search is done simultaneously with the training of networks themselves, and the search time is not significantly larger than a training time of one network.
The key idea of the one-shot NAS is the \textit{weight-sharing} trick -- that is, all the architectures from the search space share weights of architecture blocks.
%Some black-box methods could be modified by a \textit{weight-sharing} trick, that is all the architectures from the search space share weights of architecture parts. 
%Some black-box methods could be modified by a \textit{weight-sharing} 
%The \textit{weight-sharing} allows to do NAS in a one-shot manner.
Some black-box methods like evolutionary search \cite{elsken2018efficient} and RL-based NAS \cite{pham2018efficient} can be modified by the weight-sharing and enjoy considerable speedup.

The DARTS method \cite{liu2018darts} considers a supernetwork containing all the networks from a search space as its subnetworks.
The choice between subnetworks is governed by architectural parameters which are updated by gradient descent similarly to differentiable hyperparameters optimization methods \cite{pedregosa2016hyperparameter}.
%relaxes discrete search space of networks to the continuous structure of a supernetwork and optimizes its parameters similarly to differentiable hyperparameters optimization algorithms \cite{pedregosa2016hyperparameter}.
Subsequent modifications improve DARTS in terms of search speed and performance of resulting architectures \cite{xu2019pc,chen2019progressive,liang2019darts+,dong2019searching,cai2018proxylessnas}. 
Alternative approaches update subnetworks randomly during the training phase \cite{guo2019single,li2019random,bender2019understanding}.
Then, the best subnetwork is selected by its validation accuracy.
%Interestingly, subnetworks could be sampled from a supernetwork even at random \cite{guo2019single, li2019random, bender2019understanding}. These algorithms resemble dropout during the training phase since they train an exponentially large number of networks with shared weights. However, in opposite to dropout during the test phase, only the best subnetwork is selected by its validation accuracy.

%Some researches focus on detailed evaluations of existing NAS methods \cite{yang2019evaluation, adam2019understanding, li2019random}.
Overall, black-box optimization methods are much slower but more robust and general. Given a rich search space of architectures, a black-box method typically will find a good one though spending a lot of time. Also black-box optimization doesn't restrict the network's performance to be differentiable with respect to architectural parameters, like DARTS. Constraints like FLOPS/latency/memory footprint can be applied straightforwardly. 
Popular one-shot methods like DARTS and ENAS are quite fast. Unfortunately, they perform only slightly better than the random search \cite{yang2019evaluation,adam2019understanding,li2019random}.

Is it possible to speedup black-box NAS?
The natural approach is to do \textit{low-fidelity} evaluations of neural architectures, for example, train them for a few epochs.
%In this paper, we propose to speedup black-box NAS methods by using low-fidelity evaluations of neural architectures - train them on a random part of a dataset.
However, the final goal is to find the best architecture in terms of a \textit{high-fidelity} evaluation -- after training until convergence.
An interesting research question arises: is it possible to make correct architecture selection after training for a few epochs? Obviously, this selection is not perfect, but we show how to improve it by training with a \textit{knowledge distillation (KD)} loss function.
%But after training on a small part of a dataset a network loosely resembles its high-fidelity counterpart.
%To handle this issue we propose to use a \textit{knowledge distillation (KD)} loss function.
%Initially, the knowledge distillation was introduced in \cite{hinton2015distilling} for improved training of compact networks. This is done by adding to a conventional likelihood loss function a term forcing a network to mimic some teacher network. 
%Together with matching of a class label by the maximum likelihood a compact neural network tries to match predictions (softmax outputs) of another large and accurate network.
We found that the proposed technique not only improves the accuracy of a network but also improves the correlation between low- and high-fidelity evaluations.
%Finally, we show how to combine low- and high-fidelity evaluations in a \textit{multi-fidelity} search algorithm.
%In this work we propose to extend the black-box NAS methods by doing a \textit{multi-fidelity} optimization. In addition to high-fidelity evaluations of architectures, that is training on a full dataset, we do \textit{low-fidelity} evaluations by training on small parts of a dataset.

In this paper, we make the following contributions:
\begin{itemize}
    \item we propose \textbf{a new approach to the low-fidelity evaluation of neural architectures} -- training for a few epochs with a knowledge distillation loss (Section \ref{sec:method-kd});
    %\item we create and release \textbf{a new tabular benchmark for NAS}. The benchmark contains modifications of MobileNetV2 \cite{sandler2018mobilenetv2} and ShuffleNetV2 \cite{ma2018shufflenet} trained with losses of various KD methods on different subsets of the CIFAR-100 dataset (Section \ref{sec:benchmark});
    \item we incorporate the proposed low-fidelity evaluations into a \textbf{bayesian multi-fidelity search method ``\mfkd{}''} based on co-kriging (Figure \ref{fig:flowchart} and Section \ref{sec:method-mf}); % and empirically show its superiority over baselines (Section \ref{sec:experiments-mf}).
    \item we carry out experiments with the NAS-Bench-201 benchmark \cite{dong2020bench}, including CIFAR-10, CIFAR-100, and ImageNet- 16-120. We prove that the proposed approach leads to \textbf{the better architecture selection}, given the same computational budget, than several state-of-the-art baselines (Section \ref{sec:experiments});% and the naive ones -- (1) training for a few epochs with the conventional logistic loss, (2) training until convergence;
    
    %given the same computational budget.
    %: using only high-fidelity evaluations, using naive low-fidelity evaluations (training on a random subset of data).
    %\item we empirically proved that using knowledge distillation leads to selection of better neural architecture than without knowledge distillation given the same computational budget. \cmt{(figure with computational budget for KD, KD+ and no KD?)}
\end{itemize}

The code is in the repository\\ \url{https://anonymous.4open.science/r/a6c96420-435a-484b-9170-e2de9ab0aee3/}. \\

\section{Related Work}

\textbf{Knowledge distillation} (KD) was proposed in \cite{hinton2015distilling}. 
%It is a way of compressing a knowledge of large and accurate neural network (or ensemble) to a smaller network without accuracy drop.
The seminal paper matched predictions of a student and a teacher with cross-entropy.
%This is done by changing the loss function. Instead of matching class label with the maximum likelihood neural network simultaneously tries to match predictions (softmax outputs) of another network. 
Later extensions suggest matching features maps instead of class probabilities \cite{romero2014fitnets,zagoruyko2016paying,tung2019similarity,peng2019correlation,ahn2019variational,passalis2018learning,huang2017like,tian2019contrastive}.
The methods similar to KD were developed for other problems: sequences-to-sequence modeling \cite{kim2016sequence}, reinforcement learning \cite{rusu2015policy}, etc.

%Some extentions of KD include: matching neurons output from in some layer besides pre-activations of softmax \cite{luo2016face, romero2014fitnets}, application to sequences-to-sequence models \cite{kim2016sequence}, reinforcement learning \cite{rusu2015policy}.
%It is even possible to train a neural network of the same architecture on its outputs and improve an accuracy \cite{furlanello2018born}.
%Also we refer a reader to the recent review of KD methods \cite{wang2020knowledge}.

\textbf{Multi-fidelity/low-fidelity}. Low-fidelity evaluations are used sometimes in the context of hyperparameter optimization and NAS. The proposed variants include: training on a part of dataset \cite{klein2016fast}, shorter training time \cite{zela2018towards}, lower resolution of images \cite{chrabaszcz2017downsampled}, less filters per payer \cite{zoph2018learning}. Low-fidelity evaluations are faster but they are biased.
This issue motivates \textit{multi-fidelity} methods which progressively increase fidelity during the search:
%State of the art general purpose multi-fidelity optimization methods are 
MF-GP-UCB \cite{kandasamy2019multi}, MF-MI-Greedy \cite{song2018general}, co-kriging \cite{klyuchnikov2019figuring}.

\textbf{KD \& NAS}.
%The main body of knowledge distillation literature is focused on training of compact students with performance similar to a teacher or training of students outperforming a teacher.
Some very recent papers study applications of KD to NAS.
%\cite{gu2020search} proposes to find a student architecture simultaneously with the student learning. It is done by pruning of useless connections via $L_1$ regularization.
%The most similar work to ours in \cite{li2020blockwisely}.
\cite{li2020blockwisely} propose to independently train blocks in a student's supernetwork by mimicking corresponding blocks of a teacher with MSE loss. %Then an optimal combination of blocks is selected by maximizing accuracy under FLOPS and number of parameters constraints.
%In \cite{li2020blockwisely} was proposed to train sequential blocks in a student's architectures. A student's architecture is a supernetwork consisting of blocks. Each of them has several choices like convolution size and expansion ratio. Blocks of a student's supernetwork learn to imitate blocks of a teacher by minimizing MSE loss with feature maps from corresponding layers of a teacher. 
%Then an optimal architecture of a student if found - it is subnetwork of a supernetwork (particular choices for each block) with maximum accuracy under FLOPS and number of parameters constraints.
%Then an optimal subnetwork is found by maximizing accuracy under FLOPS and number of parameters constraints
\cite{kang2020towards} proposed an oracle knowledge distillation loss and showed that ENAS \cite{pham2018efficient} with this loss outperforms ENAS with logistic loss.
\cite{liu2020search} studies RL-based NAS with networks trained with KD loss instead of a logistic loss. They conclude that the found architecture depends on the teacher architecture used for KD, that is, some structural knowledge is transferred from a teacher. 

The main difference between our work and the aforementioned papers is that we use KD loss for improving low-fidelity evaluations of architectures -- inside a multi-fidelity search algorithm. At the same time, \cite{liu2020search} does only high-fidelity evaluations (training on the full dataset). \cite{kang2020towards,li2020blockwisely} incorporates KD loss into training of a supernetwork, while our work is about treating NAS as a search over a discrete domain of architectures. 
%These are different approaches to NAS.% which are hard to compare.

\section{The proposed method}
\label{sec:proposed-method}
\subsection{Knowledge Distillation (KD)}
The knowledge distillation (KD) assumes two models: a \textit{teacher} and a \textit{student}.
The teacher is typically a large and accurate network or an ensemble.
The student is trained to fit the softmax outputs of the teacher together with ground truth labels. The idea is that outputs of the teacher capture not only the information provided by ground truth labels but also the probabilities of other classes -- ``dark knowledge''. The knowledge distillation can be summarized as follows.

Let $z_i$ be logits (pre-softmax activations) and $q_i$ -- probabilities of classes as predicted by a neural network.
Knowledge distillation smooths $z_i$ with the temperature $\tau$
\begin{equation}
q_i = \sigma(z_i / \tau) = \frac{exp(z_i/\tau)}{\sum_j exp(z_j/\tau)}.
\end{equation}
Neural networks often do very confident predictions (close to 0 or 1) and smoothing helps to provide for student more information during training \cite{hinton2015distilling}. 
The KD loss is a linear combination of the logistic loss and cross-entropy between predictions of the teacher and the student
\begin{equation}
\label{kd-loss}
(1-\lambda) \sum_{i} H(y_i, \sigma(z^S_i)) + \lambda \tau^2 \sum_{i} H ( \sigma(z^T_i / \tau), \sigma(z^S_i / \tau) ),
\end{equation}
where $z^T_i$, $z^S_i$ are logits of the teacher and the student, $H(p, q) = - p \log(q)$ is the cross-entropy function. The factor $\tau^2$ is used for scaling gradients of both parts of the loss function to be the same order.
%The parameter $\lambda$ is called the \textit{imitation parameter}.
%Both predictions of the teacher $q^T_i$ are also smoothed with the temperature $\tau$.
In the rest of the paper, we will refer to this variant of the knowledge distillation as ``original KD''.

Other variants of KD suggest matching feature maps of the student and the teacher with various discrepancy functions \cite{romero2014fitnets,zagoruyko2016paying,tung2019similarity,peng2019correlation,ahn2019variational,passalis2018learning,huang2017like,tian2019contrastive}. 
For example, the NST loss \cite{huang2017like} uses Maximum Mean Discrepancy (MMD): 
\begin{equation}
\label{eq:nst-loss}
\sum_{i} \left( H(y_i, \sigma(z^S_i)) + \beta \mathcal{L}_{MMD^2}(F_{i, T}, F_{i,S}) \right),
\end{equation}
where $F_T$, $F_S$ are the feature maps of the teacher and the student, 
\begin{gather}
\mathcal{L}_{MMD^2}(F^T, F^S) = \frac{1}{C_T^2}\sum_{i=1}^{C_T}\sum_{i'=1}^{C_T} k(\frac{f_T^{i\cdot}}{||f_T^{i\cdot}||_2}, \frac{f_T^{i'\cdot}}{||f_T^{i'\cdot}||_2})
+ 
\frac{1}{C_S^2}\sum_{j=1}^{C_S}\sum_{j'=1}^{C_S} k(\frac{f_S^{j\cdot}}{||f_S^{j\cdot}||_2}, \frac{f_S^{j'\cdot}}{||f_S^{j'\cdot}||_2})\notag\\
-
\frac{2}{C_T C_S}\sum_{i=1}^{C_T}\sum_{j=1}^{C_S} k(\frac{f_T^{i\cdot}}{||f_T^{i\cdot}||_2}, \frac{f_S^{j\cdot}}{||f_S^{j\cdot}||_2})\label{eq:nst-loss-mmd}.
\end{gather}
Here $f^{i\cdot}_T$,$f^{j\cdot}_S$ are feature map from the layers $i,j$ of the teacher and the student respectively, $k(x,y)$ is a kernel.

\subsection{Low-fidelity evaluations with knowledge distillation}
\label{sec:method-kd}

Let $\alpha$ be an architecture from a search space $\mathcal{A}$. We assume that each architecture could be represented by a real-valued vector of features $x\in\mathcal{X}\subseteq\mathbb{R}^d$.
We call $y(x)$ an \textit{evaluation} of the architecture  $\alpha$, namely its validation accuracy after fitting on the train dataset. We use the following notations:
\begin{itemize}
    \item $y^1(x)$ - \textit{low-fidelity} evaluation, that is, validation accuracy of the network  $\alpha$ after fitting on the train dataset for $E_1$ epochs with the NST loss (\ref{eq:nst-loss});
    \item $y^2(x)$ - \textit{high-fidelity} evaluation, that is, validation accuracy of the network  $\alpha$ after fitting on the train dataset for $E_2$ epochs with the logistic loss, $E_2>E_1$.
\end{itemize}

The better low-fidelity evaluation $y^1(x)$ is, the higher a correlation between $y^1(x)$ and $y^2(x)$ should be.
Even when correlation is large, low-fidelity evaluations are not enough since typically they are biased: 
\begin{equation}
    \argmax_{x} y^1(x) \neq \argmax_{x} y^2(x).
    \label{eq:fidelity_opt_bias}
\end{equation}
This bias motivates \textit{multi-fidelity} search methods that combine low- and high-fidelity evaluations.
\subsection{Multi-fidelity NAS}
\label{sec:method-mf}
%In this section, we describe our algorithm for NAS with multi-fidelity search.
%The algorithm assumes that each architecture $a$ is described by a real-valued vector of features $x$. 

%The algorithm include Gaussian Process Regression (GPR) with RBF kernel \cite{williams2006gaussian} and bayesian optimization based on GPR.

We combine low-fidelily $y^1(x)$ and high-fidelity $y^2(x)$ evaluations via the co-kriging fusion model
\begin{equation}
\label{eq:co-kriging}
y^{2}(x) = \rho y^{1}(x) + \delta(x),
\end{equation}
where $y^{2}(x)$, $y^{1}(x)$, $\delta(x)$ are Gaussian Processes \cite{williams2006gaussian}, $y^{1}(x)$ is independent of $\delta(x)$, in turn, $\delta(x)$ allows handling biases \eqref{eq:fidelity_opt_bias}, $\rho$ is a scaling factor which is fitted by maximum likelihood.

Figure \ref{fig:flowchart} depicts the high-level structure of the proposed method, while Algorithm \ref{alg:mf-nas} gives the formal description.
%Algorithm \ref{alg:mf-nas} defines the proposed approach to multi-fidelity NAS with knowledge distillation.
Initially, Algorithm \ref{alg:mf-nas} samples $n_1+n_2$ architectures randomly and does low-fidelity evaluations of $n_1$ architectures and high-fidelity evaluations of $n_2$ architectures. After this warm-up, the architectures are selected cyclically by the UCB criteria (line \ref{alg_line:ucb}) and evaluated via high-fidelity evaluation only. At each iteration, the co-kriging fusion model (line \ref{alg_line:co-kriging}) is updated. Finally, Algorithm \ref{alg:mf-nas} returns an architecture having the best validation score.

%The two-fidelity model can be extended to a multi-fidelity one using the schema proposed in \cite{kennedy2000predicting}. 
The proposed model can be extended to more than two levels of fidelity using the schema proposed in \cite{kennedy2000predicting}. 
Under the Markov assumptions on the covariance structures, each level of fidelity depends only on the previous one in the same fashion as high-fidelity depends on low-fidelity in \eqref{eq:co-kriging}.

\begin{algorithm}[t!]
\caption{\mfkd{}: A Multi-fidelity Neural Architecture Search Method with Knowledge Distillation}
\label{alg:mf-nas}

\DontPrintSemicolon
\SetAlgoVlined
\SetKwInOut{Input}{Input}\SetKwInOut{Output}{Return}

\Input{$\mathcal{X}$ - search space of architectures (encodings in $\mathbb{R}^d$), $n_1$, $n_2$, $E_1$, $E_2$, $N$, T - total budget for the procedure, $\beta$ - non-negative float (default 1.0, exploration/exploitation trade-off).}
\BlankLine
$t = 0$ // spent budget\;
Randomly sample $n_{1}$ architectures -- $A_1$ \;
(I) Train architectures from $A_1$ for $E_1$ epochs with Knowledge Distillation (low-fidelity 1)\;
$t += \texttt{budget for (I)}$\;
Randomly sample $n_{2}$ architectures -- $A_2$ \;
(II) Train architectures from $A_2$ for $E_2$ epochs (low-fidelity 2)\;
$t += \texttt{budget for (II)}$\;
Fit co-kriging fusion regression 
$y^{(2)}(x) = \rho y^{(1)}(x) + \delta(x) $, where
$y^2(x)$ -- predictions for low-fidelity 2 data, 
$y^1(x)$ -- predictions for low-fidelity 1 data, 
$\delta(x)$ -- discrepancy.
$y^{(2)}(x), y^{(1)}(x), \delta(x)$ are Gaussian Processes, $y^{(1)}$ is independent of $\delta$, $\rho$ - scaling factor (parameter).\;
\While{$t < T$}
{
Sample $N$ random architectures -- $A$\;
Select one architecture $x_*$ from $A$: $x_* = \argmax_{x \in A} \left(\mathbb{E}\left[y^{(2)}(x)\right] + \beta Var\left[y^{(2)}(x)\right]\right)$ \label{alg_line:ucb}\;
(III) Train architecture $x_*$ for $E_2$ epochs (low-fidelity 2)\;
$t += \texttt{budget for (III)}$\;
Fit co-kriging fusion regression (with updated low-fidelity 2 data for $x_*$) \label{alg_line:co-kriging}
}
\BlankLine
\Output{Architecture with the best validation score after $E_2$ epochs evaluated during the procedure.}
\end{algorithm}
\section{Experiments}
\label{sec:experiments}
%In this section, we demonstrate the effectiveness of the proposed MF-KD method for neural architecture search.
%We carry out experiments using the tabular benchmark NAS-Bench-201 derived from CIFAR-10, CIFAR-100 and ImageNet-16-120 datasets.

\subsection{NAS benchmark}
\label{sec:benchmark}

\begin{figure}[t]
\centering
\includegraphics[angle=0,width=\textwidth]{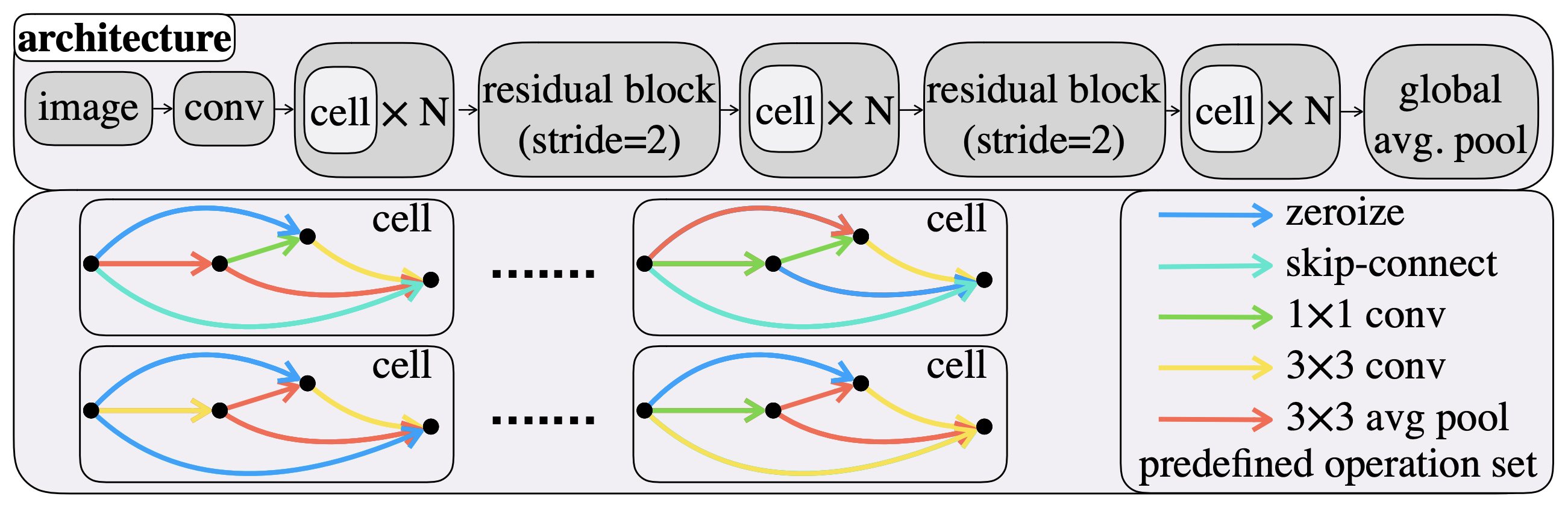}
\caption{Top: macro-structure of the network. Bottom-left: example of stacked cells with various operations. Each cell is a directed acyclic graph. Each edge is associated with some operation. Bottom-right: the list of operations. Picture was redrawn from \cite{dong2020bench}.}
\label{fig:nas-bench-201}
\end{figure}

For the experiments, we used NAS-Bench-201 \cite{dong2020bench} benchmark\footnote{Initially, we planned to carry out experiments with the larger NAS-Bench-101 \cite{ying2019bench} benchmark containing 423k trained architectures. Unfortunately, the implementation is in TensorFlow, while KD methods are implemented in PyTorch; they are not compatible. We also have faced technical difficulties with the TensorFlow code.}. It contains 15,625 convolutional architectures trained on CIFAR-10, CIFAR-100 and ImageNet-16-120 (downsampled to 16$\times$16 variant of ImageNet with 120 classes).
Figure \ref{fig:nas-bench-201} shows macro- and micro-structure of the architectures. Each cell is stacked $N = 5$ times, number of output channels gradually increases from 16 to 64 from the first to the last layer. These architectures were trained with the following hyperparameters: 200 epochs, cosine annealing learning rate, momentum 0.9, initial learning rate 0.1, weight decay $5\times10^{-4}$, and augmentation (random crop, random flip). The benchmark provides thorough logs of training.

The test accuracy of top architectures from the NAS-Bench-201 is not state-of-the-art since the networks were trained for a not so many epochs with only basic augmentation techniques. Otherwise, training of 15,625 networks would by unfeasible.

\textbf{Encoding of the architectures}. Since the macro-structure is fixed, each architecture can be unequivocally described by it's cell structure. We used concatenated one-hot encodings of operations associated with edges as encoding of the whole architecture.

%As a low-fidelity evaluation, we did training only for 1 epoch and used other values from the benchmark.

\subsection{KD methods}
\label{sec:experiments-kd}

%\begin{wraptable}{r}{6cm}
%\begin{table}[t]
%\centering
%\caption{Kendall-tau correlation between low-fidelity and high-fidelity evaluations. Low-fidelity evaluations are done by training for 1 epoch with logistic loss and NST loss.}
%\begin{tabular}{ccc}
%\toprule
%\textbf{Dataset} & \textbf{log.loss} & %\textbf{NST loss} \\
%\midrule
%CIFAR-10 & 0.17 & \textbf{0.47}\\
%CIFAR-100 & 0.06 & \textbf{0.47}\\
%ImageNet16-120 & 0.21 & \textbf{0.45}\\
%\bottomrule
%\end{tabular}
%\label{tbl:lf-hf-corr}
%\end{table}
%\end{wraptable}

\begin{table}[t]
\centering
\caption{Kendall-tau correlation between low-fidelity and high-fidelity evaluations. Low-fidelity evaluations are done by training for 1 epoch.}
\begin{tabular}{cccc}
\toprule
\textbf{Loss} & \multicolumn{3}{c}{\textbf{Dataset}} \\
\cmidrule{2-4}
 & \textbf{CIFAR-10} & \textbf{CIFAR-100} & \textbf{ImageNet16-120} \\
\midrule
logistic loss & 0.17 & 0.06 & 0.21\\
NST loss & \textbf{0.47} & \textbf{0.47} & \textbf{0.45}\\
\bottomrule
\end{tabular}
\label{tbl:lf-hf-corr}
\end{table}

In preliminary experiments with CIFAR-100 dataset (see Appendix \ref{app:kd-methods}) we tested various types of KD methods\footnote{We used implementations from \url{https://github.com/HobbitLong/RepDistiller}.}. We selected the NST method for the further experiments with NAS-Bench-201 since it lead to the highest correlation between low-fidelity and high-fidelity evaluations. In the NST loss, we used the polynomial kernel $k(x,y)=(x^Ty+c)^b$ is with $c=0,b=2$ and $\beta=12.5$ in (\ref{eq:nst-loss}). We used ResNet networks trained on the same datasets as teachers.

Unfortunately, calculating the gradient of the NST loss (\ref{eq:nst-loss}) adds significant overhead to the traditional logistic loss, training becomes $\approx$3 times slower. To mitigate this issue, we calculated the NST loss approximately
\begin{equation}
\label{eq:nst-loss-approx}
\sum_{i} \left( H(y_i, \sigma(z^S_i)) + \beta \mathcal{\widetilde{L}}_{MMD^2}(F_{i, T}, F_{i,S}) \right),
\end{equation}

with only a subset of feature maps $\widetilde{C}_T \subset \{1, \ldots, C_T\}, \widetilde{C}_S \subset \{1, \ldots, C_S\}$ \footnote{We used feature maps after 2 cells of N=5 stacked cells and each residual block (see Figure \ref{fig:nas-bench-201}).}:

\begin{gather}
\mathcal{\widetilde{L}}_{MMD^2}(F^T, F^S) = \frac{1}{|\widetilde{C}_T|^2}\sum_{i\in \widetilde{C}_T}\sum_{i' \in \widetilde{C}_T} k(\frac{f_T^{i\cdot}}{||f_T^{i\cdot}||_2}, \frac{f_T^{i'\cdot}}{||f_T^{i'\cdot}||_2})\\
+ 
\frac{1}{|\widetilde{C}_S|^2}\sum_{j \in \widetilde{C}_S}\sum_{j' \in \widetilde{C}_S} k(\frac{f_S^{j\cdot}}{||f_S^{j\cdot}||_2}, \frac{f_S^{j'\cdot}}{||f_S^{j'\cdot}||_2})\notag
-
\frac{2}{|\widetilde{C}_T| |\widetilde{C}_S|}\sum_{i \in \widetilde{C}_T}\sum_{j \in \widetilde{C}_S} k(\frac{f_T^{i\cdot}}{||f_T^{i\cdot}||_2}, \frac{f_S^{j\cdot}}{||f_S^{j\cdot}||_2})\label{eq:nst-loss-mmd-approx}.
\end{gather}

%To mitigate his issue, in (\ref{eq:nst-loss-mmd}) we calculated $MMD^2$ between only subset of feature maps, particularly after 2 cells of 5 total and each residual block (see Figure \ref{fig:nas-bench-201}). 
After this optimization, training with the NST loss became only $\approx$1.5 times slower\footnote{Even more speed-up is possible by precomputing feature maps of the teacher network.}. We didn't perform a detailed study of this issue, and consider that selecting layers for doing knowledge distillation is an interesting topic for the further research.

Table \ref{tbl:lf-hf-corr} shows Kendall-tau rank correlations between low- and high-fidelity evaluations. Low-fidelity evaluations are done by training architectures for 1 epoch. We conclude that training with the NST loss significantly improves the correlation over the conventional logistic loss.

\subsection{Multi-fidelity NAS}
\label{sec:experiments-mf}

Finally, we tested the proposed \mfkd{} method (Algorithm \ref{alg:mf-nas}) with the NAS-Bench-201 benchmark. We used parameters $E_1=1, E_2=12, n_1=100, n_2=20, N=5000$). 
NAS methods were compared by the test accuracy
\footnote{
The co-kriging regression in the Algorithm \ref{alg:mf-nas} is fitted to the \textit{validation} accuracy, while the methods are compared by the \textit{test} accuracy of the best architectures. 
Model selection based on validation accuracy while estimating performance by test accuracy is a common pattern for AutoML/NAS algorithms performed to avoid overfitting.} of the best found architecture, averaged over 100 runs. NAS methods were allowed to use the equal computational budget  -- $12\times10^3$ seconds, same as in \cite{dong2020bench}.
For all the methods except \mfkd{} we used the data from \cite{dong2020bench}.

Table \ref{tbl:mf-results} presents the results. We conclude that the \mfkd{} method is the best performing one. Particularly, it found better architecture than the state-of-the-art multi-fidelity algorithm BOHB. The improvement over the second best method, Regularized Evolution, is significant with p-value $< 0.05$.

An alternative way to assess the NAS method's quality is to compare results by the relative accuracy in the search space instead of the absolute accuracy.
The MF-KD method found architectures having accuracy's in top 0.3\%, 0.2\%, 0.5\% within the search space for the datasets CIFAR-10, CIFAR-100, and ImageNet16-120, respectively.

 %We emphasize that the networks' performances are not state-of-the-art, since experiments are limited to the architectures and training pipelines implemented in NAS-Bench-201.
%
% MULTI-FIDELITY 
%
\begin{table}[t]
\centering
\caption{Results of the \mfkd{} method. For each of the method, the accuracy of the best found architecture is shown. The search was performed under the same computational budget, averaged over 100 runs.}
\begin{tabular}{cccc}
 \toprule
 \textbf{Method} & \multicolumn{3}{c}{\textbf{Accuracy, \%}} \\
 \cmidrule{2-4}
               & \textbf{CIFAR-10} & \textbf{CIFAR-100} & \textbf{ImageNet-16-120} \\
 \midrule
 DARTS-V2 & 54.30 & 15.61 & 16.32 \\
 GDAS     & 93.51 & 70.61 & 41.84 \\
 ENAS     & 54.30 & 15.61 & 16.32 \\
 Reg. Evolution &  93.92 & 71.84 & 45.54 \\
 Random Search & 93.70 & 71.04 & 44.57 \\
 BOHB & 93.61 & 70.85 & 44.42 \\
 \textbf{\mfkd{}} & \textbf{93.93} & \textbf{72.00} & \textbf{ 45.67} \\
\bottomrule
\end{tabular}
\label{tbl:mf-results}
\end{table}

\subsection{Ablation studies}
\label{sec:ablation}
The proposed method \mfkd{} has two contributions: bayesian multi-fidelity search and low-fidelity evaluations after training with the NST loss. We carry out ablation studies of these contributions: 
\begin{itemize}
    \item Search with a single fidelity: Gaussian Processes Regression (GPR) with high-fidelity evaluations only;
    \item Multi-fidelity search, where low-fidelity evaluations are training with the conventional logistic loss for a few epochs;
\end{itemize}
Table \ref{tbl:ablation} show the results: 
both of the contributions improve the algorithm's performance for more CIFAR-100 and ImageNet-16-120 datasets. For simpler CIFAR-10 this not the case. Also, for CIFAR-100 and ImageNet-16-120 the proposed method found the architecture better than the teacher, ResNet. 

\begin{table}[t]
\centering
\caption{Ablation studies of the \mfkd{} method. For each of the method, the accuracy of the best found architecture is shown. The search was performed under the same computational budget, averaged over 100 runs.}
\begin{tabular}{cccc}
 \toprule
 \textbf{Method} & \multicolumn{3}{c}{\textbf{Accuracy, \%}} \\
\cmidrule{2-4}
               & \textbf{CIFAR-10} & \textbf{CIFAR-100} & \textbf{ImageNet-16-120} \\
\midrule
ResNet & \textbf{93.97} & 70.86 & 44.63 \\
Single fidelity (GPR)  & 93.78 & 71.44 & 45.44 \\
Multi-fidelity (no KD) & \textbf{93.97} & 71.83 & 45.41 \\
\textbf{\mfkd{}}         & 93.93 & \textbf{72.00} & \textbf{45.67} \\
\bottomrule
\end{tabular}
\label{tbl:ablation}
\end{table}
\section{Conclusion}
In this work, we have proposed the new \mfkd{} method tailored to neural architecture search.
%new approach to the low-fidelity evaluation of neural network architectures -- training for a few epochs with the knowledge distillation loss function.
By doing experiments, we have proved that the \mfkd{} method is efficient. It leads to a better architecture selection than several state-of-the-art baselines given the same computational budget. Also it outperforms state-of-the-art multi-fidelity method BOHB.
%the naive approach -- selection after training for a few epochs with the conventional logistic loss.
%The proposed low-fidelity evaluations could be applied as a standalone tool for NAS and also inside a multi-fidelity search algorithm. 
We validated our contributions on the NAS-Bench-201 benchmark, including CIFAR-10, CIFAR-100 and ImageNet-16-120 datasets. 

Our research gives an interesting insight into knowledge distillation methods themselves. While these methods are typically compared by a performance of compact student networks trained with the KD loss, we apply the KD loss to improve \textit{architecture selection} after training for a few epochs.
%We conclude that the NST \cite{huang2017like} loss function performs best is this setting.

Our work satisfies the best practices for scientific research on NAS \cite{lindauer2019best}, see Appendix \ref{appendix:best-practices}.

\bibliographystyle{splncs04}
\bibliography{kd_nas}

%
% APPENDIX
% 
%\newpage
\appendix

\section{Best practices of NAS research}
\label{appendix:best-practices}

The best practices of NAS research are the following \cite{lindauer2019best}:
\begin{enumerate}
    \item Release Code for the Training Pipeline(s) you use;
    \item Release Code for Your NAS Method;
    \item Don’t Wait Until You’ve Cleaned up the Code; That Time May Never Come;
    \item Use the Same NAS Benchmarks, not Just the Same Datasets;
    \item Run Ablation Studies;
    \item Use the Same Evaluation Protocol for the Methods Being Compared;
    \item Compare Performance over Time;
    \item Compare Against Random Sampling and Random Search;
    \item Validate The Results Several Times;
    \item Use Tabular or Surrogate Benchmarks If Possible;
    \item Control Confounding Factors;
    \item Report the Use of Hyperparameter Optimization;
    \item Report the Time for the Entire End-to-End NAS Method;
    \item Report All the Details of Your Experimental Setup.
\end{enumerate}

We released all the code (1, 2, 3). We carried out experiments on the public tabular benchmark (4, 10).
We did ablation studies in Section \ref{sec:ablation} (5).
Since we used tabular benchmark (6) is satisfied.
For multi-fidelity optimization, we made comparisons over time (7).
We compared against random search (8).
Experimental results are averaged over many runs (9).
We did our best to control confounding factors (11).
Hyperparameter optimization (12) is described in Appendix \ref{app:kd-methods}.
We did our best to report all the details about the experimental setup (14).
We also discuss computational performance in the Section \ref{sec:experiments-kd}.

\section{Comparison of KD methods}
\label{app:kd-methods}

Initially, we evaluated various KD methods on two small search spaces: 100 modifications of MobileNetV2 \cite{sandler2018mobilenetv2} and 300 modifications of ShuffleNetV2 \cite{ma2018shufflenet} architectures.

These architectures share the same pattern -- particular human-designed blocks are repeated certain number of times while channels count increase from input to output.
To create search spaces, we randomly modified repetitions and channels counts while preserving the increasing pattern of channels from input to output. These numbers -- repetitions and channels count -- were used as architectures' features. The dimensionality of the MobileNetV2 search space is 16, the ShuffleNetV2 search space -- 7.
To avoid too small and too large architectures, we left only ones having a number of parameters in the range $(\nicefrac{1}{3}P, 3P)$, where $P$ is the number of the parameters of the original MobileNetV2 and ShuffleNetV2 respectively. 
We have trained all the architectures on the full CIFAR-100 \cite{krizhevsky2009learning} dataset, and its \nicefrac{1}{27}, \nicefrac{1}{9}, \nicefrac{1}{3} random but fixed subsets (instead of few epochs).
Various loss functions were tested: logistic loss (no KD), knowledge distillation methods:  original KD \cite{hinton2015distilling}, Hint \cite{romero2014fitnets}, AT \cite{zagoruyko2016paying}, SP \cite{tung2019similarity}, CC \cite{peng2019correlation}, VID \cite{ahn2019variational},
PKT \cite{passalis2018learning}, NST \cite{huang2017like},
CRD \cite{tian2019contrastive}. 
The hyperparameters of training were: 100 epochs, momentum 0.9, cosine annealing learning rate, initial learning rate 0.1, weight decay $5\times10^{-4}$, batch size 128 with random cropping and horizontal flipping. The hardware was GeForce GTX 1080 Ti.
Teachers in search spaces were original MobileNetV2 and ShuffleNetV2 architectures trained on the same dataset with the same hyperparameters.

%\subsection{Search spaces \& hyperparameters}
%\label{sec:search-spaces}
%{\color{red}TODO: refactoring}.
%
% PEARSON CORR.
%
\begin{table}[t]
\centering
\caption{Correlation between high-fidelity and low-fidelity evaluations.
}
\label{tbl:many-kd-corr}
\centering
 \begin{tabular}{cccccccccccc}
 \toprule
 \multirow{2}{*}{\textbf{Part}} & \multicolumn{10}{c}{\textbf{Pearson corr.}} \\
 \cmidrule{2-11} & \textbf{no KD} & \textbf{orig. KD} & \textbf{AT} & \textbf{NST} & \textbf{SS} & \textbf{VID} & \textbf{PKT} & \textbf{CRD} & \textbf{Hint} & \textbf{CC}\\
 \midrule
      & \multicolumn{10}{c}{{MobileNetV2 search space}}\\
 \cmidrule{2-11}
 1/27 & 0.11 & 0.34 & \textbf{0.57} & 0.42 & 0.35 & -0.03 & 0.35 & 0.18 & 0.19 & 0.16\\
 %\midrule
 1/9 & 0.46 & \textbf{0.61} & \textbf{0.61} & 0.60 & 0.53 & 0.07 & 0.47 & 0.44 & 0.48 & 0.45\\ 
 %\midrule
 1/3 & 0.86 & \textbf{0.92} & 0.74 & 0.81 & 0.79  & -0.21 & 0.41 & 0.85 & 0.84 & 0.90\\ 
 \midrule
      & \multicolumn{10}{c}{{ShuffleNetV2 search space}}\\
 \cmidrule{2-11}
 1/27 & 0.48 & 0.54 & 0.43 & \textbf{0.61} & 0.45 & 0.45 & 0.44 & 0.43 & 0.47 & 0.46\\
 %\midrule
 1/9 & 0.64 & \textbf{0.81} & 0.57 & 0.74 & 0.61 & 0.60 & 0.60 & 0.30 & 0.64 & 0.58\\ 
 %\midrule
 1/3 & \textbf{0.92} & 0.91 & 0.72 & \textbf{0.93} & 0.91 & 0.92 & 0.76 & 0.88 & 0.91 & \textbf{0.92}\\ 
 \bottomrule
\end{tabular}
\end{table}

%\section{Optimal hyperparameters of KD methods}
%\label{app:hyperparams}
\textbf{Hyperparameters tuning}.
We have tuned hyperparameters of KD methods by doing low-fidelity evaluations of 20 random architectures with training on a \nicefrac{1}{9} part of the CIFAR-100 dataset. Then we have selected the best combination by the highest correlation with high-fidelity evaluations, see Table \ref{tbl:kd-hyperparams}. The same hyperparameters were used for the NST loss for main experiments with NAS-Bench-201 benchmark (Section \ref{sec:experiments}). 
Table \ref{tbl:many-kd-corr} shows correlation between high-fidelity and low-fidelity evaluations for the best hyperparameters. We conclude that the AT and NST loss (AT loss is the particular case of NST loss) perform the best for the evaluation by training on \nicefrac{1}{27} of the dataset. 

\begin{table}[t]
    \centering
    \caption{Optimal hyperparameters of KD methods}
    \label{tbl:kd-hyperparams}
    \begin{tabular}{lcc}
        \toprule
        KD method & MobileNetV2 & ShuffleNetV2 \\
                  & search space & search space \\
        \midrule
        Distilling the knowledge in a neural network
        \cite{hinton2015distilling} (KD) & \multicolumn{2}{c}{$\tau=32, \lambda=1$} \\
        Fitnets: Hints for thin deep nets \cite{romero2014fitnets} (Hint) & \multicolumn{2}{c}{$\beta=100$} \\
        Attention Transfer (AT) \cite{zagoruyko2016paying} & $\beta=10^3$ & $\beta=4\times10^3$ \\
        Similarity-Preserving Knowledge Distillation (SP) \cite{tung2019similarity} & $\beta=750$ & $\beta=90$ \\
        Correlation Congruence (CC) \cite{peng2019correlation} & \multicolumn{2}{c}{$\beta=0.5\times10^{-2}$} \\ 
        Variational information distillation &  $\beta = 0.01$ &  $\beta=0.25$ \\
        \quad for knowledge transfer (VID) & & \\
        Learning deep representations  & \multicolumn{2}{c}{$\beta=48\times10^4$}  \\
        \quad with probabilistic knowledge transfer (PKT) \cite{passalis2018learning} & & \\
        Like what you like: & $\beta=12.5$ & $\beta=200$ \\
        \quad Knowledge distill via neuron select. transfer (NST) \cite{huang2017like} & & \\
        Contrastive Representation Distillation (CRD) \cite{tian2019contrastive} & $\tau=0.2, \beta=0.5$ & $\tau=0.05, \beta=1$\\
        \bottomrule
    \end{tabular}
\end{table}

%The original KD loss is described in (\ref{kd-loss}), losses of other KD methods in (\ref{eq:inner-kd}), where always $\alpha=1$.

\section{Additional experiments with ImageNet}
\label{app:imagenet}
We did low-fidelity evaluations of all the architectures from the MobileNetV2 search space by training on \nicefrac{1}{27} part of the ImageNet dataset. For low-fidelity evaluations, we trained for 100 epochs and other hyperparameters were the same as for low-fidelity evaluations on CIFAR-100.

For high-fidelity evaluations, we used the following hyperparameters: 150 epochs with momentum 0.9, cosine annealing learning rate, initial learning rate 0.05, weight decay $4\times10^{-5}$, batch size 128, the augmentation included random cropping and horizontal flipping. 

Additionally, we did high-fidelity and low-fidelity evaluations of 10 random architectures and calculated Kendall-tau correlation between high- and low-fidelity evaluations. For conventional logistic loss, it turned out 0.42, while for the original KD loss 0.73. The increase in the correlation confirms our conclusions.
\end{document}